\documentclass{article} 
\usepackage[final]{colm2025_xllmworkshop}

\usepackage{microtype}
\usepackage{hyperref}
\usepackage{url}
\usepackage{booktabs}

\usepackage{algorithm}
\usepackage{algpseudocode}
\usepackage{amsmath}
\usepackage{amsthm}
\usepackage{amsfonts}

\usepackage{tcolorbox}
\usepackage{verbatim}

\definecolor{darkblue}{rgb}{0, 0, 0.5}
\hypersetup{colorlinks=true, citecolor=darkblue, linkcolor=darkblue, urlcolor=darkblue}

\tcbset{
promptstyle/.style={
before skip=1.5em,  
after skip=1.5em,
boxsep=2pt,
top=5pt,
bottom=5pt
}
}

\title{Before you \texttt{<think>}, monitor:\\Implementing Flavell's metacognitive framework in LLMs}

\author{Nick Oh \\
socius labs\\
London, UK\\
\texttt{nick.sh.oh@socius.org}
}

\begin{document}

\ifcolmsubmission
\linenumbers
\fi

\maketitle

\begin{abstract}
Current approaches to enhancing LLM reasoning follows two isolated paradigms: \textit{Monitor-Generate} methods like Plan-and-Solve \citep{wang2023plan} and SELF-DISCOVER \citep{zhou2024self} excel at strategic planning but lack mechanisms to verify whether selected strategies succeed; while \textit{Generate-Verify} approaches like Self-Verification \citep{weng2022large} and SELF-REFINE \citep{madaan2023self} iteratively refine outputs but commence generation blindly without task assessment. This separation creates inefficiencies -- strategies fail without feedback, and refinement occurs without strategic grounding. We address this gap by implementing \citeauthor{flavell1979metacognition}'s cognitive monitoring model (\citeyear{flavell1979metacognition}) from the broader \textit{Monitor-Generate-Verify} framework \citep{oh2025mgv}, operationalising it as a three-phase iterative system. On GSM8K, preliminary results show 75.42\% accuracy versus 68.44\% for SELF-REFINE and 67.07\% for Self-Verification, while requiring fewer attempts (1.3 vs 2.0) at 27-37\% increased inference cost. These initial findings suggest upfront monitoring produces higher-quality initial solutions that reduce refinement needs, though evaluation beyond arithmetic reasoning is needed to establish generalisability.
\end{abstract}

\section{Introduction}

This preliminary experimental report documents our initial implementation and testing of \citeauthor{flavell1979metacognition}'s (\citeyear{flavell1979metacognition}) cognitive monitoring model for enhancing LLM. We operationalise the \textit{Monitor-Generate-Verify} framework proposed by \citet{oh2025mgv}, creating a system where models explicitly assess task difficulty before attempting solutions, adapt their computational resources accordingly, and evaluate outputs along structured metacognitive dimensions. Our experiments on a subset of GSM8K problems compare this approach against Self-Verification and SELF-REFINE baselines using \texttt{Llama-3.1-8B-Instruct}. While our evaluation is limited to 659 arithmetic problems with a single model, the results suggest that explicit monitoring may reduce the need for extensive iterative refinement by producing better initial solution attempts. We present these findings as an early exploration of whether cognitive science principles, particularly metacognitive monitoring, can inform more effective reasoning architectures for language models.

This paper is structured as follows: Section~\ref{sec:relatedworks} situates our approach within existing paradigms. Section~\ref{sec:mgv} presents our implementation of Flavell's cognitive monitoring model. Section~\ref{sec:experiments} evaluates performance against baselines. Section~\ref{sec:limitations} discusses limitations and future directions. Section~\ref{sec:discussion} reflects on the methodological implications of our approach.

\section{Related Works}
\label{sec:relatedworks}

\subsection{Monitor-Generate (MG) Methods}

\textit{Monitor-Generate} (MG) approaches recognise that effective reasoning requires understanding task structure before attempting solutions. \textit{Zero-shot prompting methods} have evolved from simple planning to sophisticated strategy discovery: Plan-and-Solve \citep{wang2023plan} replaces generic ``think step by step'' with structured guidance for problem decomposition and careful execution, achieving 76.7\% accuracy on arithmetic tasks; SELF-DISCOVER \citep{zhou2024self} enables LLMs to identify task-intrinsic reasoning structures through meta-prompts (SELECT, ADAPT, IMPLEMENT), yielding 27-32\% improvements on reasoning tasks; and Meta-Reasoning Prompting \citep{gao2024meta} functions as an adaptive router, evaluating multiple reasoning methods against task requirements to select the optimal approach. \textit{Few-shot prompting methods} leverage exemplars for enhanced strategy selection, with Strategic Chain-of-Thought \citep{wang2024strategic} first eliciting strategic knowledge then applying it systematically (achieving 21.05\% improvement on GSM8K), and HYBRIDMIND \citep{han2024hybridmind} employing a meta-selector to choose between natural language, symbolic, or hybrid reasoning. \textit{Reinforcement learning based approaches} like Elastic Reasoning (Scalable Chain-of-Thought) \citep{xu2025scalable} optimise strategy through trained budget allocation between thinking and solution phases. Despite their sophistication in pre-generation planning, these methods lack mechanisms to verify whether selected strategies actually succeed or to learn from failed attempts.

\subsection{Generate-Verify (GV) Methods}

\textit{Generate-Verify} (GV) approaches focus on iterative refinement through self-evaluation. Self-Verification \citep{weng2022large} generates multiple candidates then validates through backward reasoning, systematically masking conditions to reconstruct original problems and selecting based on aggregated verification scores. SELF-REFINE \citep{madaan2023self} implements a three-phase cycle where the same model serves as generator, critic, and refiner, maintaining complete iteration history to prevent repeating mistakes, achieving 20\% average improvement without additional training. Recent advances emphasise sophisticated verification mechanisms: ReVISE \citep{lee2025revise} enables intrinsic self-verification through structured curriculum learning with confidence-aware decoding, while theoretical work formalises self-improvement through the `generation-verification gap' that scales with model pre-training FLOPs \citep{song2024mind}. The pivotal role of verifier quality has been demonstrated by \citet{zhang2024small}, showing substantial gains with strong verifiers but limitations with weak self-verifiers. However, even strong verification cannot fully overcome the `prefix dominance trap' \citep{luo2025learning}: once models begin with suboptimal reasoning strategies, performance degrades by nearly 20\% with minimal recovery through subsequent verification. This exposes a fundamental limitation -- these approaches commence generation without assessing task characteristics or retrieving problem-solving strategies, leading to inefficient exploration that initial strategic planning could avoid.

\section{\texttt{Monitor-Generate-Verify} Framework}
\label{sec:mgv}

The preceding analysis reveals a fundamental gap. MG methods excel at strategic planning but cannot verify success, while GV approaches refine outputs iteratively but lack initial task assessment. Addressing these complementary limitations, we implement the \textit{Monitor-Generate-Verify} (MGV) framework \citep{oh2025mgv} -- specifically operationalising \citeauthor{flavell1979metacognition}'s cognitive monitoring model (\citeyear{flavell1979metacognition}) as a three-phase iterative reasoning system. Rather than operating within a single paradigm, MGV synthesises both: it monitors task characteristics to inform strategy selection (addressing GV's blind generation), executes with adaptive parameters, then verifies outcomes to guide subsequent cycles (addressing MG's lack of systematic validation). This creates a complete metacognitive loop where verification informs monitoring, monitoring guides generation, and generation produces solutions whose evaluation feeds back to refine initial assessments.

The framework operates through $T$ cycles, where each cycle $\tau \in \{0, 1, ..., T-1\}$ consists of three phases. Our implementation employs \textit{zero-shot prompting} throughout, avoiding the need for task-specific examples while testing the generality of metacognitive principles (complete prompts are provided in Appendix~\ref{app:prompts}). Following Flavell's assumption of pre-existing metacognitive knowledge, we adopt the 20 problem-solving strategies compiled by \citet{didolkar2024metacognitive}, who instructed LLMs to assign skill labels to GSM8K problems. While this knowledge base was constructed using \texttt{gpt-4-0613}, raising questions about cross-model transferability -- akin to humans having privileged access to their own metacognition \citep{nelson1990metamemory} and findings that LLMs better predict their own behaviour than other models' \citep{binder2024looking} -- we adopt it as a reasonable starting point for this preliminary investigation. This knowledge base primarily represents $\mathcal{MK}_{\text{Strategy}}$ from Flavell's three categories ($\mathcal{MK}_{\text{Agent}}$, $\mathcal{MK}_{\text{Task}}$, $\mathcal{MK}_{\text{Strategy}}$), focusing on problem-solving approaches rather than agent capabilities or task characteristics.

\begin{algorithm}
\caption{\texttt{Flavell's Model of Cognitive Monitoring}}
\begin{algorithmic}[1]
\Require task $\mathcal{T}$, model $M$, strategies $\mathcal{MK}$, prompts $\mathcal{P}=\{p_{monitor}, p_{strategy}, p_{execute}, p_{verify}\}$
\State Initialize $\mathcal{S}_0 = \text{ACTIVE}$, $\tau = 0$, $\mathcal{ME}_{\text{evaluative}}^{-1} = \emptyset$
\While{$\mathcal{S}_\tau = \text{ACTIVE}$ and $\tau < T$}
\State \textbf{// Monitor: Assess task difficulty}
\State $\mathcal{ME}_{\text{difficulty}}^\tau, features_\tau = M(p_{mon} \| \mathcal{T} \| \mathcal{ME}_{\text{evaluative}}^{\tau-1})$ \Comment{Assess task difficulty}
\State
\State \textbf{// Generate: Apply cognitive strategy}
\State $strategy_\tau = M(p_{str} \| features_\tau \| \mathcal{ME}_{\text{difficulty}}^\tau \| \mathcal{MK})$ \Comment{Choose approach}
\State $solution_\tau = M(p_{exe} \| \mathcal{T} \| strategy_\tau)$ \Comment{Execute strategy with adaptive params}
\State
\State \textbf{// Verify: Evaluate performance}
\State $\mathcal{ME}_{\text{evaluative}}^\tau = M(p_{ver} \| \mathcal{T} \| solution_\tau)$ \Comment{Evaluate output quality}
\State
\State $\mathcal{S}_{\tau+1} = \textbf{if } \text{mean}(\mathcal{ME}_{\text{evaluative}}^\tau) \geq 0.85 \textbf{ then } \text{TERMINATE } \textbf{else } \text{ACTIVE}$
\State $\tau = \tau + 1$
\EndWhile
\State \Return $y_{\arg\max_i \text{mean}(\mathcal{ME}_{\text{evaluative}}^i)}$
\end{algorithmic}
\end{algorithm}

The following subsections detail each component of our implementation, examining how monitoring assess task difficulty, how generation employs dual-stage processing, and how verification provides structured metacognitive feedback. 

\subsection{Monitor}

The model analyses problems without solving them, identifying task characteristics and assessing difficulty on a [0,1] scale (Appendix~\ref{app:mgv_monitor}). While \citeauthor{steyvers2025metacognition} (\citeyear{steyvers2025metacognition}) propose \textit{implicit elicitation} of metacognitive states through token likelihood measurements, we employ \textit{explicit elicitation}, prompting the model to verbalise its difficulty assessment. For $t > 0$, the monitor receives evaluation scores from the previous cycle $(coherence, plausibility, consistency, goal\_conduciveness) \in [0,1]^4$, enabling metacognitive recalibration -- lower scores trigger upward difficulty adjustment, which subsequently increases computational resources during generation.

This abstraction mechanism, where detailed failure information is compressed into difficulty reassessment, speculatively parallels human metacognition: rather than explicitly recalling every step of failed attempts, we often carry forward an implicit sense that ``this is harder than I thought'' that shapes subsequent approaches. Setting aside the cognitive parallel, this compression offers a computationally efficient alternative to maintaining complete episodic traces, encoding failure patterns directly into resource allocation decisions.

\subsection{Generate}

Generation employs a two-stage approach: strategy selection from 20 domain-specific approaches based on monitored features and difficulty, followed by solution execution with adaptive parameters (Appendices~\ref{app:mgv_generate}). Harder problems receive proportionally more resources -- token budgets scale as $400 + \mathcal{ME}_{\text{difficulty}} \times 400$ and temperature as $0.3 + \mathcal{ME}_{\text{difficulty}} \times 0.2$ -- reflecting that harder problems may benefit from expanded exploration while maintaining sufficient determinism for reliable computation. For $t > 0$, generation receives complete previous cycle context (attempted strategy, previous reasoning, diagnostic feedback), preventing redundant approaches and enabling targeted improvements.

The cognitive plausibility of separating selection from execution is well-established. \citet{reder1987strategy} identifies a distinct ``strategy selection phase'' in human question answering preceding execution, while dual strategy models \citep{beeson2019mental} emphasise selection as a separate cognitive act. Empirical evidence reinforces this distinction. \citet{rickard2004strategy} demonstrates that practice transitions people from multi-step algorithmic strategies to direct retrieval, highlighting selection-execution independence. Research on arithmetic problem-solving further shows these functions are independently affected by cognitive load \citep{imbo2007working}.

\subsection{Verify} 

Verification evaluates solutions along four Flavellian dimensions -- coherence (logical connectivity), plausibility (approach reasonableness), consistency (computational accuracy), and goal-conduciveness (question answering) -- producing both numerical scores $\in [0,1]^4$ and diagnostic text explaining specific strengths or failures (Appendix~\ref{app:mgv_verify}). The framework terminates when mean evaluation score $\geq 0.85$ or after $T$ cycles, implementing a satisficing strategy that accepts good-enough solutions rather than pursuing optimality. This structured evaluation distinguishes strategy selection errors (requiring different approaches) from execution errors (requiring careful reapplication), informing the monitoring phase's subsequent recalibration and creating the complete metacognitive loop.

\section{Experimental Results}
\label{sec:experiments}

We evaluate our implementation of Flavell's MGV model against Self-Verification and SELF-REFINE baselines (Appendix \ref{app:appendix_algorithms}) on 659 randomly sampled problems from the GSM8K test set \citep{cobbe2021training}. Comparing across this spectrum from simple verification of multiple candidate solutions to iterative refinement offers preliminary insights into whether gains stem from monitoring or merely additional computation.

\subsection{Experimental Setup}

All experiments utilised \texttt{Llama-3.1-8B-Instruct} with task-specific zero-shot prompt configurations on an NVIDIA H100 SXM GPU. For MGV, we implement Flavell's framework with $T=3$ maximum cycles, terminating early if the mean evaluation score exceeds 0.85. The Monitor phase assesses difficulty $\in [0,1]$, which dynamically adjusts generation parameters: token budget scales as $400 + 400 \times \mathcal{ME}_{\text{difficulty}}$ and temperature as $0.3 + 0.2 \times \mathcal{ME}_{\text{difficulty}}$. Strategy selection draws from 20 predefined approaches compiled by \citet{didolkar2024metacognitive}. The Verify phase evaluates solutions along four dimensions (coherence, plausibility, consistency, goal-conduciveness), each scored on [0,1] with 300 tokens allocated for evaluation.

Self-Verification generates 3 candidate solutions at temperature 0.3 (800 tokens each), then validates through backward verification with majority voting at temperature 0.3. SELF-REFINE iterates through generate-critique-refine cycles with temperature 0.3 (800 tokens), terminating after 3 cycles or upon receiving positive feedback. 

\subsection{Results}

Table 1 presents comparative results across accuracy, computational cost, and solution attempts. Here, "attempts" indicate reasoning cycles before termination: for MGV, each attempt is one Monitor-Generate-Verify cycle; for SELF-REFINE, each attempt is one generate-critique-refine iteration; for Self-Verification, attempts represent candidate generation and verification rounds using majority voting. 

\begin{table}[h]
\centering
\label{tab:results}
\begin{tabular}{lccc}
\toprule
\textbf{Method} & \textbf{Accuracy} & \textbf{Avg Time (s)} & \textbf{Avg Attempts} \\
\midrule
Self-Verification & 442/659 (67.07\%) & 7.52 & 1.2 \\
Self-REFINE & 451/659 (68.44\%) & 6.98 & 2.0 \\
\textbf{MGV (Flavell)} & \textbf{497/659 (75.42\%)} & 9.60 & 1.3 \\
\bottomrule
\end{tabular}
\caption{Performance comparison on GSM8K (659 problems)}
\end{table}

Flavell's MGV model achieves 75.42\% accuracy on GSM8K, a 7-8 percentage point improvement over both baselines and representing a 22\% relative error reduction compared to SELF-REFINE. Notably, this gain is achieved with fewer average attempts than SELF-REFINE (1.3 vs 2.0), with approximately 70\% of problems solved on the first cycle -- suggesting that metacognitive monitoring produces higher-quality initial solutions that reduce the need for extensive iteration. This efficiency in attempt count tentatively validates that explicit monitoring helps avoid the prefix dominance trap \citep{luo2025learning}, where poor initial strategies cascade into unrecoverable errors. However, these benefits incur computational overhead: MGV requires 27.7\% more time than Self-Verification and 37.5\% more than SELF-REFINE, adding approximately 2-3 seconds per problem due to the monitoring and strategy selection phases. This trade-off -- higher accuracy with fewer attempts but increased inference time -- positions MGV as particularly suitable for applications prioritising solution quality over real-time constraints, while potentially prohibitive for latency-critical deployments.

\section{Limitations and Future Directions}
\label{sec:limitations}

Beyond the immediate limitations of evaluating solely on arithmetic reasoning and lacking multiple-run statistical validation, our work exposes several key questions.

\paragraph{Probing Metacognitive Monitoring} A fundamental limitation lies in our reliance on \textit{explicit elicitation} \citep{steyvers2025metacognition} -- prompting models to verbalise their metacognitive experience of difficulty. These outputs depend on the model's ability to represent and articulate internal states in language, which may not accurately reflect actual computational processes. Studies consistently find that implicit confidence measures derived from token likelihoods exhibit greater metacognitive sensitivity than explicitly prompted confidence \citep{xiong2023can}, highlighting a gap between what models internally represent and what they express. This discrepancy is particularly concerning given \citet{lindsey2025biology}'s demonstration that while \texttt{Claude-3.5-Haiku} accurately reports intermediate steps, it hallucinates non-existent computational processes when describing its internal mechanisms for simple addition, despite correctly activating relevant neural pathways. More promising directions involve implicit estimation of metacognitive states. Research reveals that features like confidence and certainty correspond to linearly separable directions in representation space \citep{zou2023representation, liu2023aligning}. Notably, \citet{ji2025language} identify a low-dimensional `metacognitive space' within LLMs' neural representations, suggesting models monitor a subset of their mechanisms -- opening possibilities for direct neural probes of metacognitive experiences including Feeling-of-Knowing, Ease-of-Learning, and Judgements-of-Learning in future MGV implementations.

\paragraph{Modular Architecture and Verification Challenges} Our uniform 8B model overlooks differential capability requirements across phases. \citet{qin2025decomposing} decompose mathematical reasoning into Plan, Execute, and Verify -- strikingly parallel to MGV framework -- revealing that small models (0.5B) reliably identify solution plans while execution demands substantially larger models, suggesting potential efficiency gains from size-differentiated architectures. However, verification poses the greatest challenge. \citet{stechly2024self} demonstrate self-critique consistently degrades performance through false negatives, hallucinated feedback, and insensitivity to critique levels, with improvements emerging only through external sound verifiers. Notably, performance gains prove independent of critique richness -- multiple attempts with reliable verification suffices -- aligning with \citet{zhang2024small}'s findings on verifier criticality and explaining Self-Verification's efficiency (1.2 attempts through candidate generation). MGV's design -- abstracting feedback into difficulty rather than specific corrections, and switching strategies rather than editing solutions -- may reduce error propagation, though self-verification remains fundamentally unreliable. Future implementations should explore modular architectures: smaller monitoring models (3B), standard generation models (8B), and either substantially larger models or external symbolic verifiers, creating an LLM-Modulo framework \citep{kambhampati2024llms} leveraging specialised capabilities while maintaining metacognitive coherence.

\paragraph{Inherent Reasoning Boundaries} \citet{qin2025decomposing} observe that execution failures often stem from models applying steps based on spurious correlations -- superficial formatting or phrasing patterns -- rather than robust reasoning, echoing the prefix dominance trap \citep{luo2025learning}. While reinforcement learning improves execution by reducing low-level errors, it cannot expand fundamental mathematical understanding, only optimising what models already implicitly know. This suggests MGV shares inherent limits in addressing genuinely novel problems beyond the model's latent capabilities. The framework enhances navigation of known solution spaces but cannot transcend the underlying model's reasoning boundaries.

\paragraph{Metacognitive Knowledge and Adaptation} Adopting \texttt{gpt-4-0613}'s strategies \citep{didolkar2024metacognitive} for \texttt{Llama-3.1-8B} neglects potential model-specific metacognitive representations, paralleling how humans possess privileged access to their own metacognition \citep{nelson1990metamemory}. Future work should explore learning model-intrinsic metacognitive knowledge through self-supervised methods. The Monitor component could leverage V-STaR's approach \citep{hosseini2024vstar}, which utilises both successful and failed reasoning attempts -- training via preference learning to distinguish effective from ineffective strategies rather than discarding failures. Applied to MGV, this would enable the Monitor to learn from strategy selection errors: contrasting high-scoring solutions against low-scoring ones to progressively refine difficulty assessments and strategy mappings. Through iterative bootstrapping across reasoning cycles, the system could discover model-specific problem taxonomies and emergent strategies beyond those available in external knowledge bases, transforming MGV from imposing borrowed metacognitive structure to cultivating genuine model-intrinsic metacognitive capabilities.

\section{Discussion}
\label{sec:discussion}

MGV is not technically ``novel'' in its individual computational mechanisms. Its monitoring phase -- assessing task difficulty and selecting strategies -- parallels the pre-generation analysis found in MG paradigm. And its generate-verify loop with iterative refinement mirrors GV paradigm. What might seem distinctive -- combining upfront strategic planning with iterative refinement -- can be approximated by sequentially applying existing methods. Nevertheless, the four evaluation dimensions (coherence, plausibility, consistency, goal-conduciveness), while structured, function as evaluation metrics rather than fundamental algorithmic innovations. From a purely technical perspective, MGV appears to be a well-engineered integration of established techniques.

Yet the distinction lies not in technical superiority but in methodological origins. While MG and GV methods draw \textit{inspiration} from human reasoning processes, using cognitive insights to guide engineering decisions, MGV tests whether formal psychological theories can directly \textit{translate} into computational systems. This theory-first implementation's preliminary success opens an underexplored research direction: established cognitive theories can become testable machine learning hypotheses. This approach could potentially accelerate progress by leveraging decades of theoretical and empirical work rather than rediscovering patterns through engineering iteration. Both paths -- inspiration and implementation -- may enrich our understanding of how reasoning principles might transcend their original substrates.

\bibliography{colm2025_conference}
\bibliographystyle{colm2025_conference}

\appendix

\newpage
\section{MGV (Monitor-Generate-Verify) Zero-shot Prompting Templates}
\label{app:prompts}

We present the complete prompting templates used in our experiments. Variables are denoted in curly brackets \texttt{\{variable\}}.

\subsection{Monitor Phase}
\label{app:mgv_monitor}
\begin{tcolorbox}[title=\texttt{Monitor}, promptstyle]
\begin{verbatim}
Problem: {problem}
{if t > 0:
evaluation_scores:
coherence: {score}
plausibility: {score}
consistency: {score}
goal_conduciveness: {score}}

Before you <think>, analyse this problem WITHOUT solving it.

Output format:
<monitor>
Task_Features: List 2-3 keywords describing the math concepts needed
Difficulty (0-1): Assess the challenge level
{if t > 0: Recalibrate your assessment based on the
evaluation_scores in previous_cycle. Lower scores
suggest higher actual difficulty than previously assessed.}
</monitor>
\end{verbatim}
\end{tcolorbox}

\newpage
\subsection{Generate Phase}
\label{app:mgv_generate}
\begin{tcolorbox}[title=\texttt{Generate - Strategy Selection}, promptstyle]
\begin{verbatim}
Problem: {problem}
Task features: {task_features}
Difficulty: {difficulty:.3f}
Available strategies: {comma_separated_strategy_list}

Select ONE strategy from the list above that best fits this problem.

Output format:
Selected Strategy: [exact strategy name from list]
\end{verbatim}
\end{tcolorbox}

\texttt{Available strategies = [multiplication and addition, basic arithmetic, addition and multiplication, arithmetic operations, multiplication, percentage calculations, subtraction, algebra, subtraction and division, multiplication and division, multiplication and subtraction, addition and subtraction, percentage calculation, addition subtraction, average calculation, subtraction multiplication, division, addition, linear equations, algebraic reasoning]}.

\begin{tcolorbox}[title=\texttt{Generate - Strategy Execution}, promptstyle]
\begin{verbatim}
Problem: {problem}
Strategy: {strategy_type}

Output format:
<think>
Work through the problem step by step using the {strategy_type} approach.
Show all calculations and reasoning.
{if t > 0:
previous_cycle:
difficulty: {difficulty}
task_features: {task_features}
strategy_used: {strategy_type}
reasoning: [full solution text]
evaluation_scores:
coherence: {score}
plausibility: {score}
consistency: {score}
goal_conduciveness: {score}
evaluation: [full evaluation text]}
</think>

<answer>
Output only the final numerical answer (no units or text).
</answer>
\end{verbatim}
\end{tcolorbox}

\subsection{Verify Phase}
\label{app:mgv_verify}
\begin{tcolorbox}[title=\texttt{Verify}, promptstyle]
\begin{verbatim}
Problem: {problem}
Solution: {solution}

Evaluate this solution on these dimensions (0-1 scale):

1. COHERENCE: Do the steps logically connect?
2. PLAUSIBILITY: Is the approach reasonable?
3. CONSISTENCY: Are calculations correct?
4. GOAL-CONDUCIVENESS: Does it answer the question?

Output format:
<evaluate>
Coherence: X.X
Plausibility: X.X
Consistency: X.X
Goal-conduciveness: X.X

Evaluation: [Provide a 2-3 sentence analysis explaining the scores.
Identify specific errors or strengths. For low scores, indicate
what went wrong (e.g., "arithmetic error in step 3", "misunderstood
the question", "skipped crucial reasoning"). For high scores, note
what worked well. Be specific and actionable.]
</evaluate>
\end{verbatim}
\end{tcolorbox}

\newpage
\section{Baseline Algorithm Implementations}
\label{app:appendix_algorithms}

\begin{algorithm}
\caption{Self-Verification algorithm \citep{weng2022large}}
\begin{algorithmic}[1]
\Require input $x$, model $M$, prompt $p_{cot}$, candidates $K$, verifications $P$
\State $\{y_1, \ldots, y_K\} = M(p_{cot} \| x)$ with sampling \Comment{Generate diverse answers}
\For{each candidate $y_k$}
\State $score_k = 0$
\State $conclusion_k = \text{Rewrite}(x, y_k)$ \Comment{Turn Q\&A into statement}
\For{$p = 1$ to $P$} \Comment{Verify multiple times}
\For{each condition $c_i$ in $x$}
\State $x_{masked} = \text{Replace}(x, c_i, \text{"X"}) + conclusion_k$
\State $\hat{c_i} = M(p_{cot} \| x_{masked} \| \text{"Find X"})$ \Comment{Backward reasoning}
\If{$\hat{c_i} = c_i$}
\State $score_k = score_k + 1$ \Comment{Successful reconstruction}
\EndIf
\EndFor
\EndFor
\EndFor
\State \Return $\arg\max_{k}(score_k)$ \Comment{Most verifiable answer wins}
\end{algorithmic}
\end{algorithm}

\begin{algorithm}
\caption{SELF-REFINE algorithm \citep{madaan2023self}}
\begin{algorithmic}[1]
\Require input $x$, model $M$, prompts $\{p_{gen}, p_{fb}, p_{refine}\}$, stop condition $\text{stop}(\cdot)$
\State $y_0 = M(p_{gen} \| x)$ \Comment{Initial generation (Eqn. 1)}
\For{iteration $t \in 0, 1, \ldots$}
\State $fb_t = M(p_{fb} \| x \| y_t)$ \Comment{Feedback (Eqn. 2)}
\If{$\text{stop}(fb_t, t)$}
\State \textbf{break} \Comment{Stop condition}
\Else
\State $y_{t+1} = M(p_{refine} \| x \| y_0 \| fb_0 \| \ldots \| y_t \| fb_t)$ \Comment{Refine (Eqn. 4)}
\EndIf
\EndFor
\State \Return $y_t$
\end{algorithmic}
\end{algorithm}

\end{document}